\documentclass{article}
\usepackage{spconf,amsmath,graphicx,booktabs,multirow}
\usepackage[colorlinks,linkcolor=black,anchorcolor=black,citecolor=black,urlcolor=blue]{hyperref}

\title{TrOMR:Transformer-based Polyphonic Optical Music Recognition}
\name{Yixuan Li, Huaping Liu, Qiang Jin\sthanks{Corresponding author: \href{mailto:jinqiang@corp.netease.com}{jinqiang@corp.netease.com}}, Miaomiao Cai, Peng Li}

\address{NetEase Cloud Music, Hangzhou, China}
\begin{document}
%
\maketitle
\begin{abstract}
Optical Music Recognition (OMR) is an important technology in music and has been researched for a long time. Previous approaches for OMR are usually based on CNN for image understanding and RNN for music symbol classification. In this paper, we propose a transformer-based approach with excellent global perceptual capability for end-to-end polyphonic OMR, called TrOMR. We also introduce a novel consistency loss function and a reasonable approach for data annotation to improve recognition accuracy for complex music scores. Extensive experiments demonstrate that TrOMR outperforms current OMR methods, especially in real-world scenarios.  We also develop a TrOMR system and build a camera scene dataset for full-page music scores in real-world. The code and datasets will be made available for reproducibility\footnote[1]{The code and datasets will be made available at: \url{https://github.com/NetEase/Polyphonic-TrOMR}}.
\end{abstract}

\section{Introduction}
\label{sec:intro}

Optical Music Recognition (OMR), which is the foundation for digitization and intelligence of music, has great potential in various applications, such as propagation of digital music materials and retrieval of music information\cite{VictorArroyo2022NeuralAM,PedroRamoneda2022ScoreDA}. Although OMR has been studied for decades, it is still a challenge to recognize music symbols with perfect precision, especially in real-world scenarios.

\begin{figure}[htb]
\centering 
\includegraphics[width=0.475\textwidth]{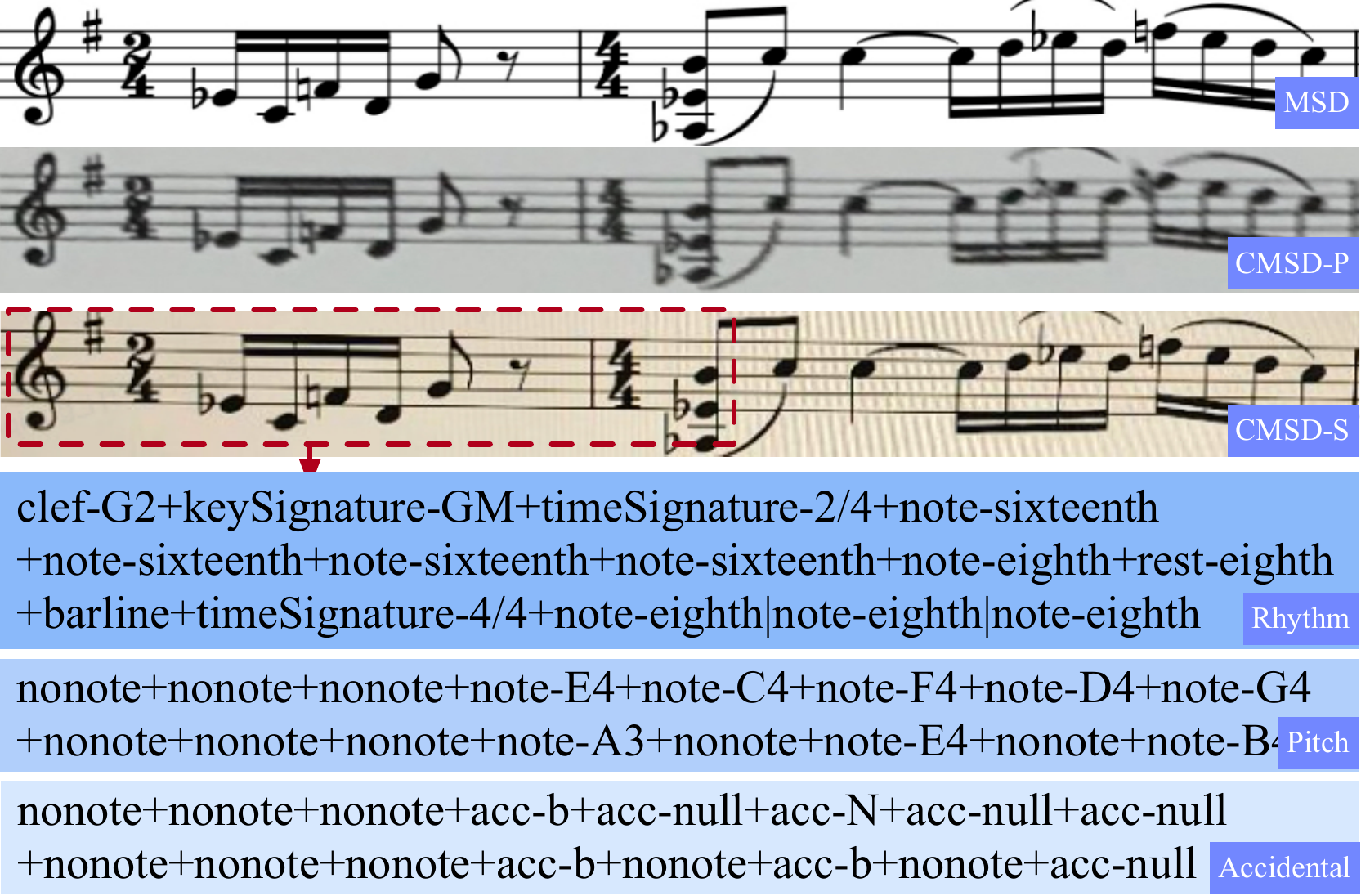}
\caption{Samples of MSD and CMSD. Rows from top to bottom represent printed staff, paper-photo staff, screen-photo staff, rhythm label, pitch label, and accidental label.}
\label{Fig.dataset} 
\end{figure}

Traditional OMR approaches typically use a multi-stage strategy consisting of image preprocessing\cite{JorgeCalvoZaragoza2017PixelwiseBO,QuangNhatVo2016AnMM}, music object detection\cite{AliciaForns2011TheI2,ZhiqingHuang2019StateoftheArtMF,AlexanderPacha2018ABF} and music symbol reconstruction\cite{ChristopherRaphael2011NEWAT,JanHaji2018TowardsFH}. Nevertheless, the traditional OMR framework usually suffers from complex processing steps, and its performance is still not satisfactory. In recent years, driven by deep learning, the end-to-end OMR methods have been widely studied\cite{EelcovanderWel2017OpticalMR,MaraAlfaroContreras2019ApproachingEO,PrIMuSNE}. The mainstream of these approaches usually follow the classical Convolutional Recurrent Neural Network(CRNN) with a variety of training losses, such as Connectionist Temporal Classification (CTC) loss, cross-entropy loss, and Smooth-L1 loss. By taking into account the spatial relationships and long-term dependencies between symbols, they achieve better performance on music scores.

\begin{figure*}[htb]
\centering 
\includegraphics[width=1.0\textwidth]{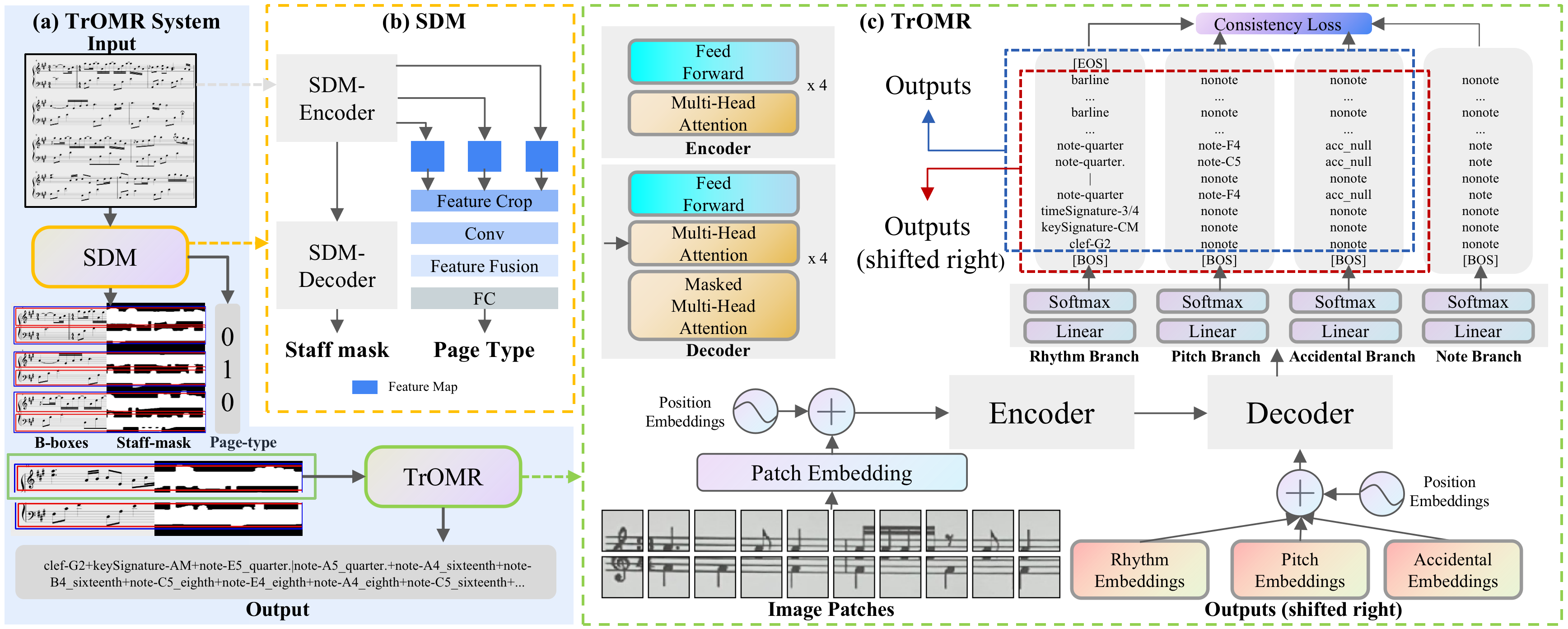}
\caption{Illustrations of the TrOMR system, SDM, and TrOMR in this paper.}
\label{Fig.network} 
\end{figure*}

However, the existing end-to-end OMR approaches, like\cite{PrIMuSNE,CameraPrIMuSNE} and \cite{MaraAlfaroContreras2019ApproachingEO}, tend to work in both monophonic and homophonic music\cite{ElonaShatri2020OpticalMR}, while few research on polyphonic music is studied. To tackle the complex polyphony, a two-branch RNNdecoder model\cite{Baseline} is proposed which can predict the pitch and rhythm of the music symbols at the horizontal position for each single vertical image slice. This method, without information interaction between the branches, is confronted with the problem of semantic mismatch of symbols located at the same position between the two predicted sequences, and it also lacks of generality for the data distributed differently from the training set, such as music score images in real world.

In this paper, motivated by the extraordinary ability of transformer to perceive contextual information\cite{LongshenOu2022ExploringTP,MinghaoLi2021TrOCRTO,AlexeyDosovitskiy2020AnII}, we propose TrOMR, an efficient transformer-based approach to end-to-end OMR. Our TrOMR has three modules: patch embedding, encoder and decoder. The patch embedding first map staff image into a D-dimensional patch embedding vector. The encoder following the ViT-style model\cite{AlexeyDosovitskiy2020AnII} takes the embedded feature vectors as input. Finally, the decoder predicts the corresponding music symbols. We also design a consistency loss function, which can significantly reduce mismatches of outputs, to strengthen the semantic consistency of sequences. Comparing to \cite{Baseline}, a reasonable approach for data annotation is performed to improve recognition accuracy for our TrOMR.

Considering practicality, we develop a TrOMR system, consisting of a staff detection module (SDM) and TrOMR, for full-page music score recognition and build a Camera MuseScore Dataset (CMSD) for generalization capability. SDM analyzes the layout of the full-page music score and detects all single staves, and TrOMR recognizes the music symbols on the detected staves.

\section{MUSIC SCORE DATASET}
\label{sec:datasets}
Several datasets have been proposed for end-to-end OMR\cite{PrIMuSNE, Baseline, CameraPrIMuSNE}, while these datasets are usually different from real-word scenarios. In this section, we present a dataset of photographs and a method of data annotation.

\subsection{Data Generation}
\label{ssec:data_generation}
Following the method in [18], we generate about 0.4 million single staff images from MuseScore files with a ratio of 4/6 between polyphonic and non-polyphonic images, called MuseScore Dataset (MSD). Further, we present a Camera MuseScore Dataset (CMSD) consisting of three sub-datasets, CMSD-Page, CMSD-P and CMSD-S. CMSD-Page is created by randomly selecting 10,000 full-page images from above MuseScore files, while half of these images are displayed on the screen and the others are printed on paper, and taking photos with mobile phone under variety conditions such as bright and dark environment, distortion and blur. Then we crop these images to 71,292 single staff images to generate screen-photo dataset(CMSD-S) and paper-photo dataset(CMSD-P). Fig.\ref{Fig.dataset} shows an example of our CMSD.

\subsection{Data Annotation}
\label{ssec:data_annotation}

In this paper, a symbol sequence by semantic encoding is used to represent a single staff, similar to \cite{Baseline}. Considering the difficulty of pitch prediction as demonstrated in \cite{Baseline} and the extensibility of the symbols, we decouple the semantic encoding into three parts, i.e., rhythm, pitch(ignore accidental) and accidental, as shown in Fig.\ref{Fig.dataset}. Rhythm sequence represents the rhythm of the notes and the non-note symbols (e.g. clefs, key signatures, time signatures and barlines). The chord notes,  from bottom to top, in the rhythm sequence are linked by "\textbar". Pitch and accidental sequence represent note attributes, and they add a "nonote" label representing the non-note symbol to align with the rhythm sequence. This annotation method makes our alphabet smaller and extension of more symbols easier in the future.

\section{METHODOLOGY}
\label{sec:methodology}
We propose TrOMR with a consistency loss to efficiently tackle end-to-end OMR task and a system that composes of SDM and TrOMR to handle full-page score images. The overall architecture of TrOMR system is shown in Fig.\ref{Fig.network}.

\subsection{Staff Detection Module}
\label{sec:SDM}
Since music scores are generally not regular that some symbols are small or extend beyond the staff, such as notes on ledger lines. In this section we use semantic segmentation and full-page label to detect the staves on a full-page score. In particular, we precisely compute a bounding box that covers all symbols associated with the staff.

As shown in Fig.\ref{Fig.network}, given an input score image, following the encoder-decoder method\cite{LiangChiehChen2018EncoderDecoderWA}, a staff mask is generated, and it considers the staves and music symbols as foreground and others as background. At the same time, our SDM directly predicts a full-page type label $C$ to analyze the layout of the input score image. Empirically, we usually analyze the layout according to the accolade which is on the left side of the staff. Then we cut out the left quarter of the feature maps in the encoder to predict $C$. In this 
work, $C$ is defined as three classes, one-staff system, two-staves system, and others.

After obtaining staff mask and $C$, a accurate bounding box for single staff region is computed by horizontal projection method. The full page-type $C$ is used to combine the results of TrOMR for staves in the same accolade.

\subsection{TrOMR}
\label{ssec:TrOMR}
TrOMR inherits the encoder-decoder structure of Transformer. The encoder extracts the visual features of the staff patches, and the decoder generates the staff symbol sequence by combining the features of encoder and the previous predictions.

The input of the encoder is a single staff image, which is resized according to the predefined height $H$ while maintaining the aspect ratio, and is disassembled into $N= HW /P^2$ patches with the fixed size $(P, P)$. Then we use ResNet to extract the feature vectors and map them into D-dimensional patch embedding vectors by linear projection. The learnable one-dimensional positional embedding is then added to the patch embedding.

We adopt the original Transformer decoder\cite{AshishVaswani2017AttentionIA} as the decoder of TrOMR, as shown in Fig.\ref{Fig.network}. The output of decoder is directed to four branches, i.e., Rhythm branch, Pitch branch, Accidental branch, and Note branch. The first three branches are designed to predict the corresponding sequence as described in Section 2.2, and the Note branch is responsible for determining the symbol on the staff is a note or a non-note. In the following, the above four branches are represented by $B_1\sim B_4$.

In the training phase, $t_1\sim t_3$ represent rhythm tokens, pitch tokens, and accidental tokens, and $PE$ represents the decoder position embedding. We add the right-shifted embeddings $Embedding(t_1^i)\sim Embedding(t_3^i)$ to $PE$ as the input $z$ of decoder, as shown in equation (\ref{con:decoderinput}).

\begin{equation}
    z^i=\sum_{c=1}^3Embedding(t_c^i)+ {PE}^i
\label{con:decoderinput}
\end{equation}

The main effect of the Note branch is to strengthen the model's ability of extracting note semantics and further improve recognition accuracy. It is also used to calculate the consistency loss with the other three branches to reduce inconsistencies between outputs.

We try to use the output of the first, second, and last hidden layer in the decoder as input of the note branch. The experiments show that using the output of the last hidden layer as input of the note branch has the best effect for training.

\begin{table}[t]
    \centering
    \caption{Results(SER) of architecture comparison }
    \label{Table.architecture}
\begin{tabular}{lcccc}
\hline
TestSet                     & Architecture & Pitch  & Rhythm & Merge \\ \hline
\multirow{3}{*}{MSD-poly}   & baseline\cite{Baseline}     & 0.035  & 0.028  & 0.127   \\
                            & TrOMR\_A & 0.028  & 0.021  & 0.069   \\
                            & \textbf{TrOMR}        & \textbf{0.021}  & \textbf{0.015}  & \textbf{0.025}   \\ \hline
\multirow{3}{*}{MSD-nonpoly} & baseline\cite{Baseline}     & 0.009  & 0.008  & 0.294   \\
                            & TrOMR\_A & 0.008  & 0.008  & 0.017   \\
                            & \textbf{TrOMR}        & \textbf{0.004} & \textbf{0.003} & \textbf{0.005}   \\ \hline
\end{tabular}
\end{table}

\subsection{Consistency Loss Function}
\label{ssec:obfun}

As shown in Equation (\ref{con:lossfunc}), our overall loss function is composed of two parts: cross- entropy loss $L_{ce}$ and consistency loss $L_{con}$. In our experiments, $\lambda=0.1$ and $\beta=1.0$.

The output of each branch $B_c(z)$ with softmax activation is first used to calculate the cross-entropy loss. In computing the consistency loss, we first accumulate $B_1(z)\sim B_3(z)$ to the two-dimensional vectors $B^{'}_1(z) \sim B^{'}_3(z)$, depending on the label belongs to note or non-note. As the consistency standard, $B_4(z)$ is used to compute the L1 norm with $B^{'}_1(z) \sim B^{'}_3(z)$ respectively. The calculation formula is shown in Equation (\ref{con:consist}). The consistency loss constrains the output symbols of all branches into the same category. 

\begin{equation}
    L_{TrOMR}=\lambda L_{ce} + \beta L_{con}
\label{con:lossfunc}
\end{equation}

\begin{equation}
    L_{con}=\frac{\sum_{c=1}^3\Vert B^{'}_c(z)-B_4(z) \Vert_1}{3}
\label{con:consist}
\end{equation}

\section{EXPERIMENTS}
\label{sec:experiments}
In this section, we present several experiments to validate the performance of our architecture for the OMR task and the effectiveness of the proposed dataset.

\subsection{Experiment Setup}
\label{ssec:setup}

We implemented all our models using Pytorch and trained them on a single A100 GPU with the memory of 40GB. The batch size is set to 32, and ADAM optimizer is used with an initial learning rate = 0.001.

For TrOMR, we use a $128 \times 1280$ image and $16 \times 16$ patches for the encoder. Both encoder and decoder have 4 layers, 256 hidden sizes and 8 attention heads. We randomly select 10,814 samples of the same content from MSD and CMSD as testing sets respectively, and the rest as training sets. Symbol Error Rate (SER) is used as the evaluation metric, described as Equation (\ref{con:ser}), where $I,S,D$ is the number of insertions, substitutions and deletions, and $N$ is the number of symbols in the reference sequence. Our experimental results will be reported from three directions: rhythm, pitch and merge rhythm and pitch.

\begin{equation}
	SER=\frac{S+D+I}{N}
\label{con:ser}
\end{equation}

\subsection{Architecture Comparison}
\label{ssec:architecture}

We investigate the effectiveness of the proposed architecture through extensive experiments and compare to architecture with RNNDecoder in \cite{Baseline} which is represented as the baseline in this paper. To ensure fairness, TrOMR is modified with a two-branches transformer decoder which resembles baseline, called TrOMR\_A. The pitch and rhythm labels of baseline are used for TrOMR\_A. Baseline model was fine-tuned with pre-trained weights, and all experiments are performed on MSD.

As illustrated in Table \ref{Table.architecture}, TrOMR\_A performs better than baseline, especially for polyphonic music, which indicate that the transformer can ameliorate contextual feature extraction process. Compared to baseline and TrOMR\_A, TrOMR achieves the best results. We argue that this is due to our data annotation and consistency loss method can also be used to improve the performance of TrOMR, and more experiments will be conducted in the next sections.

\begin{table}[t]
    \centering
    \caption{Results(SER) of consistency experiments}
    \label{Table.consistency}
\begin{tabular}
{p{0.17\textwidth} p{0.07\textwidth} p{0.07\textwidth} p{0.07\textwidth}}
\hline
\multirow{2}{*}{Architecture} & \multicolumn{3}{c}{MSD}  \\
                              & Pitch & Rhythm & Merge \\ \hline
TrOMR\_B            & 0.026 & 0.023  & 0.047   \\
TrOMR\_C         & 0.007 & 0.005  & 0.008   \\
\textbf{TrOMR}          & \textbf{0.006} & \textbf{0.004}  & \textbf{0.007}   \\ \hline
\end{tabular}
\end{table}

\subsection{Consistency Experiments}
\label{ssec:branch}
In this section, we conduct a series of ablation experiments to investigate the impact of the consistency loss. In Table \ref{Table.consistency}, TrOMR\_B represents TrOMR with no consistency loss, and TrOMR\_C represents  TrOMR with the rhythm branch as consistency standard.

As reported, we notice that the merge-SER is higher than rhythm-SER and pitch-SER in all experiments, indicating that there are some inconsistency between pitch and rhythm sequences. The performance of the TrOMR\_B are significantly suppressed compared to the TrOMR\_C and TrOMR, which verify the effectiveness of the consistency loss. Also, the distances between the merge-SER and the pitch-SER, rhythm-SER of TrOMR\_C and TrOMR are significantly smaller, indicating that the predicted pitch and rhythm sequence have a higher consistency. In addition, TrOMR has a lower SER than TrOMR\_C, suggesting that an independent note branch can achieve a better effect.

\begin{table}[t]
\centering
\caption{Results(SER) of generalization experiments}
\label{Table.generalization}
\begin{tabular}{lcccc}
\hline
TestSet                      & Architecture    & Pitch & Phythm & Merge \\ \hline
\multirow{3}{*}{CMSD-P}  & baseline(M)   & 0.881 & 0.969  & 1.162   \\
                                & TROMR(M)      & 0.066 & 0.060  & 0.081   \\
                                & \textbf{TROMR(M+C)} & \textbf{0.022} & \textbf{0.019}  & \textbf{0.024}   \\ \hline
\multirow{3}{*}{CMSD-S} & baseline(M)   & 0.912 & 0.935  & 0.968   \\
                                & TROMR(M)      & 0.467 & 0.479  & 0.528   \\
                                & \textbf{TROMR(M+C)} & \textbf{0.022} & \textbf{0.021}  & \textbf{0.023}   \\ \hline
CMSD-P+S & TROMR(M+C)   & 0.022 & 0.020  & 0.024   \\
CMSD-page& TROMR System   & 0.024 & 0.020 & 0.027   \\ \hline

\end{tabular}
\end{table}

\subsection{Generalization Experiments}            
\label{ssec:generalization}
We use baseline and TrOMR on the CMSD for generalization experiments and show the results in the top two rows of Table \ref{Table.generalization}. The training dataset is marked in parentheses, where M is MSD and C is CMSD. When only MSD is used for training, TrOMR performs better than baseline in all testing sets. In addition, SER is higher on CMSD-S than CMSD-P, which is due to the fact that most images in CMSD-S have moire pattern. TrOMR trained on both MSD and CMSD achieves the best results which means that our dataset can improve the ability of our model in real-world scenarios.

Next, we conduct experiments to evaluate the performance of SDM. The input image of SDM is blurred and scaled to $256 \times 1024$. We assess the SDM with precision rate and recall rate by Intersection over Union (IoU) with a threthold of 0.4. For the AudioLabs v2 dataset\cite{2019_ZalkowVTAM_MeasureAnnotation_ISMIR-LBD}, 860 one-staff and two-staves music score images are selected, containing 9,978 bounding boxes for staff annotations, the precision rate and recall rate are $98.45\%$ and $98.75\%$. On our CMSD-page, we obtained a precision rate of $98.99\%$ and a recall rate of $97.32\%$.

The recognition results for the TrOMR system on the CMSD-page are shown in the last row of Table\ref{Table.generalization} , which are only slightly worse than those of TrOMR on CMSD-P and CMSD-S, demonstrating its remarkable ability to handle complex scores in real-world scenarios.


\section{CONCLUSIONS}
\label{sec:Conclusions}

In this work, we propose a transformer-based architecture for polyphonic optical music recognition and create a Camera MuseScore Dataset by manually photographing for real-world OMR task. Moreover, we introduce the Staff Detection Module to develop a full-page TrOMR system. The experimental results show that our method achieves state-of-the-art performance and generalization. In future works, we will keep on
researching more efficient data augmentation strategies and expanding more musical symbols such as dynamics, ties, and tuplets to build a general TrOMR system.

\bibliographystyle{IEEEbib}
\bibliography{strings,refs}

\end{document}